\pdfoutput=1

\documentclass[11pt]{article}

\usepackage[preprint]{acl}

\usepackage{times}
\usepackage{latexsym}

\usepackage[T1]{fontenc}

\usepackage[utf8]{inputenc}

\usepackage{microtype}

\usepackage{inconsolata}
\usepackage{xspace}
\newcommand{\modelname}[0]{{\textsc{Svpo}}\xspace}

\usepackage{amsmath}
\usepackage{amsfonts}
\usepackage{graphicx}
\usepackage{booktabs}
\usepackage{multirow}
\usepackage{pifont}
\usepackage{fancyvrb}
\usepackage{subcaption}
\usepackage{colortbl}
\usepackage{comment}

\definecolor{darkgreen}{RGB}{50,100,0}
\definecolor{darkred}{RGB}{200, 0, 0}
\newcommand{\cmark}{\textcolor{darkgreen}{\ding{51}}} %
\newcommand{\xmark}{\textcolor{darkred}{\ding{55}}}

\usepackage[most]{tcolorbox}
\definecolor{wkgreen}{RGB}{184,244,175}
\definecolor{wkpurple}{RGB}{210,210,253}
\definecolor{wkyellow}{RGB}{255,241,177}
\definecolor{wkblue}{RGB}{174,217,253}

\usepackage{algpseudocodex}
\usepackage{algorithm}
\usepackage{hyperref}

\title{Step-level Value Preference Optimization for Mathematical Reasoning}

\author{%
  Guoxin Chen, Minpeng Liao, Chengxi Li, Kai Fan\thanks{Corresponding Author.}\\
  Tongyi Lab \\
  \texttt{chenguoxin22@mails.ucas.ac.cn}\\
  \texttt{\{minpeng.lmp,xiji.lcx,k.fan\}@alibaba-inc.com} \\
}

\begin{document}
\maketitle
\begin{abstract}
Direct Preference Optimization (DPO) using an implicit reward model has proven to be an effective alternative to reinforcement learning from human feedback (RLHF) for fine-tuning preference aligned large language models (LLMs). 
However, the overall preference annotations of responses do not fully capture the fine-grained quality of model outputs in complex multi-step reasoning tasks, such as mathematical reasoning. 
To address this limitation, we introduce a novel algorithm called \textit{Step-level Value Preference Optimization} (\modelname). 
Our approach employs Monte Carlo Tree Search (MCTS) to automatically annotate step-level preferences for multi-step reasoning. 
Furthermore, from the perspective of learning-to-rank, we train an explicit value model to replicate the behavior of the implicit reward model, complementing standard preference optimization. 
This value model enables the LLM to generate higher reward responses with minimal cost during inference. 
Experimental results demonstrate that our method achieves state-of-the-art performance on both in-domain and out-of-domain mathematical reasoning benchmarks.
Our code is available at \url{https://github.com/MARIO-Math-Reasoning/Super_MARIO}.
\end{abstract}

\section{Introduction}

Recently, large language models (LLMs) have demonstrated remarkable capability across a wide range of natural language processing (NLP) tasks~\citep{openai2023gpt4,du-etal-2022-glm,team2023gemini,chen-etal-2023-mprompt,anil2023palm,bai2023qwen,llamma3blog}. 
However, they continue to encounter significant challenges when engaging in complex and symbolic multi-step reasoning, particularly in mathematical reasoning~\citep{chen2022program,azerbayev2023llemma,yu2023metamath,shao2024deepseekmath,chen-etal-2024-seer,kang2024mindstar,chen2024alphamath}.

Most existing studies~\citep{wang2023mathcoder,yue2023mammoth,gou2023tora,lu2024mathgenie,liao2024mario} have significantly improved the mathematical reasoning capabilities through fine-tuning on high-quality positive supervision data (\emph{i.e}, correct solutions) annotated by GPT-4. 
In this process, a large number of negative examples generated by GPT-4 are wasted, and the model blindly imitates successful cases without understanding what the wrong solutions are.
Therefore, preference learning, such as Direct Preference Optimization (DPO)~\citep{RafailovSMMEF23}, has been proposed to align with human preferences and enable the model to distinguish between positive and negative examples. 
However, most current efforts~\citep{yuan2024self,chen2024self,pang2024iterative} focus on solution-level preferences, relying on humans or GPT-4 to generate and score complete solutions for training. 
This approach is expensive and often provides only a coarse preference relationship, which does not reflect the natural process by which humans learn to solve mathematical problems.
This discrepancy arises because solution-level preferences pursue a solution to its final answer, without informing which step in the negative solution (\emph{i.e.,} $\mathbf{y}^{l}$) led to the mistake.
Unlike these approaches, humans tend to identify and analyze their mistakes step by step when learning to solve mathematical problems, thereby preventing repeated errors.
In this manner, humans progressively learn to make informed decisions in similar states.

\begin{table}[t]
\centering
    \resizebox{0.95\linewidth}{!}{
\begin{tabular}{@{}ccccc@{}}
\toprule
\multirow{2}{*}{\begin{tabular}[c]{@{}c@{}}Training\\ Paradigm\end{tabular}} & \multicolumn{2}{c}{Training Data} & \multirow{2}{*}{\begin{tabular}[c]{@{}c@{}}Annotation\\ from GPT-4\end{tabular}} & \multirow{2}{*}{Preference} \\ \cmidrule(lr){2-3}
                                                                             & Pos.            & Neg.            &                                                                                  &                             \\ \midrule
SFT                                                                          & \cmark          & \xmark          & \cmark                                                                           & \xmark                      \\
DPO                                                                          & \cmark          & \cmark          & \cmark                                                                           & Solution-level              \\ \midrule
\modelname (Ours)                                                                         & \cmark          & \cmark          & \xmark                                                                           & Step-level                  \\ \bottomrule
\end{tabular}
}
\caption{
Comparison of different training paradigm.
}
\label{tab:method_comparison}
\vspace{-0.2cm}
\end{table}

Furthermore, while DPO reparameterizes the reward function in reinforcement learning from human feedback (RLHF)~\citep{Ouyang0JAWMZASR22} to improve simplicity and training stability, it also discards the state-value function $V(\mathbf{s})$, which is used to evaluate the expected return from the current state.
Recent work~\citep{abs-2309-15028,liao2024mario} has demonstrated the effectiveness of the value model in improving the reasoning capabilities of policy models, but it is limited by the need for additional annotated data or the complexity of the reinforcement learning process.

\begin{figure*}[t]
    \centering
    \includegraphics[width=1.0\linewidth]{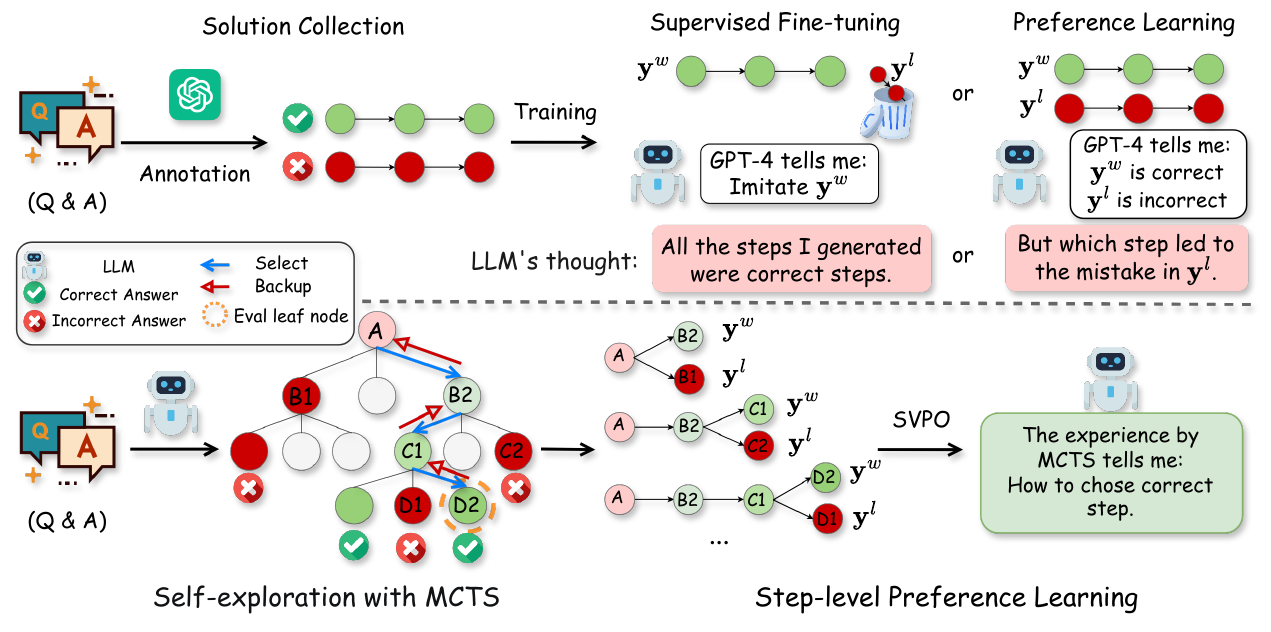}
    \caption{Comparison of different frameworks: SFT, DPO, and \modelname. The top panel shows the typical pipeline of SFT and DPO, where GPT-4 does not indicate which step in $\mathbf{y}^l$ led to the mistake. The bottom panel illustrates the pipeline of \modelname. Step-level preferences are autonomously generated via MCTS, where $Q$-values (represented by node colors) indicate potential reasoning errors.}
    \label{fig:pipeline}
\end{figure*}

To address the above issues, we propose \textit{Step-level Value Preference Optimization} (\modelname), a novel preference learning framework that focuses on more fine-grained step-level preferences via Monte Carlo Tree Search (MCTS) to significantly enhance mathematical reasoning capabilities.
Specifically, as illustrated in Figure~\ref{fig:pipeline}, step-level preferences are autonomously generated through the MCTS framework~\citep{SilverHMGSDSAPL16,SilverSSAHGHBLB17}. This approach not only avoids labor-intensive annotation but also provides detailed insights into which steps may lead to mistakes in $\mathbf{y}^l$, as indicated by the $Q$-value at each node.
Compared to the forced knowledge infusion through GPT-4 annotated data, the preferences obtained through self-exploration are better aligned with the capabilities of the current LLM, highlighting the reasoning errors that the model is more prone to making.
Furthermore, we integrate an explicit value model with DPO, where the value model is designed not only to assist the policy model (\emph{i.e.}, LLM) in navigating more effective reasoning paths but also to steer preference learning.
In our work, the value model is trained based on both $Q$-values and step-level preference relationships derived from MCTS, thereby bypassing the need for additional annotations and simplifying the training process.

We conduct extensive experiments on both in-domain and out-of-domain mathematical reasoning datasets. Our \modelname significantly outperforms state-of-the-art methods, achieving comparable or even superior results to GPT-4 on 7B LLMs.
The experiments demonstrate three key points:
\textbf{(1)}, the self-exploration process via MCTS naturally provides step-level preference relationships and highlights potential reasoning errors by $Q$-values;
\textbf{(2)}, compared to solution-level preferences, step-level preferences can significantly enhance the reasoning capabilities of the policy model;
\textbf{(3)}, the value model effectively guides the policy model's preference learning and reasoning.

\section{Background}

In standard RLHF framework, it first learns a reward model $r(\mathbf{x}, \mathbf{y})$ with Bradley-Terry~\citep{bradley1952rank} preference optimization.
\begin{equation}
\small
\mathcal{L}(r) = -\mathbb{E}_{\mathbf{x}, \mathbf{y}^w, \mathbf{y}^l} \left[ \log\sigma\left(r(\mathbf{x}, \mathbf{y}^w) - r(\mathbf{x}, \mathbf{y}^l) \right) \right]
\end{equation}
where the expectation is taken over a preference dataset that includes tuples of prompts and preference responses $(\mathbf{x}, \mathbf{y}^w \succ \mathbf{y}^l)$. 
Following this, the policy model $\pi$ is optimized using the learned reward model $r$ and the proximal policy optimization (PPO) algorithm~\citep{SchulmanWDRK17}. 
Typically, PPO requires maintaining 4 models in the training pipeline: a reward model $r$, a policy model $\pi$, a reference policy model $\pi^\prime$, and a value model $V$, making it a complex procedure.

Instead of learning an explicit reward model, DPO only maintains 2 policy models and minimize the following objective.
\begin{equation}
\small
\mathcal{L}_{\text{DPO}}(\pi) = - \mathbb{E}_{\mathbf{x}, \mathbf{y}^w, \mathbf{y}^l} \left[ \log \sigma\left( \beta \log\frac{\pi (\mathbf{y}^w|\mathbf{x}) \pi^\prime (\mathbf{y}^l|\mathbf{x})}{\pi^\prime (\mathbf{y}^w|\mathbf{x})\pi (\mathbf{y}^l|\mathbf{x})} \right) \right]  \label{eq:dpo_loss}
\end{equation}
where reference policy $\pi^\prime$ is typically a supervised fine-tuning (SFT) model. 
The implicit reward model is characterized by the log-likelihood ratio between two policy models.
Although DPO simplifies the training process, it discards the value model, which has been proven effective in improving the reasoning capabilities of the policy model.
Additionally, the coarse preferences derived from existing annotation methods also limit its performance in multi-step reasoning tasks.

\section{Method}
In this section, we present our \modelname in detail to further explore the potential of preference learning in multi-step reasoning tasks, particularly in mathematical reasoning.

\subsection{Step-level Preference Annotation}
\label{sec:mcts}

Unlike the traditional annotations that only provide solution-level preferences, we employ the MCTS framework to encourage LLMs to autonomously explore step-level generations as well as infer step-level preferences, as shown in Figure~\ref{fig:pipeline}.
In this manner of self-exploration, we obtain more fine-grained preferences, while the $Q$-values at each step (tree node) indicate potential reasoning errors that traditional annotations cannot achieve. 

The policy model for running MCTS is a SFT model of multi-step reasoning, denoted as $\pi(\mathbf{y}_{1:T}|\mathbf{x})$, where $\mathbf{x}$ is the prompted question and $\mathbf{y}_t$ represents the $t$-th step. 
In the parlance of reinforcement learning, the state and action are defined as $\mathbf{s}_t=\mathbf{y}_{<t}$ and $\mathbf{a}_t = \mathbf{y}_t$, respectively. 
In addition, the state transition function is deterministic as $\mathbf{s}_{t+1} = \text{ConCat}[\mathbf{s}_t, \mathbf{a}_t]$.

Our primary objective in annotating preferences is to compare the quality of two potential step-level generations. 
Concretely, this can be transformed into comparing the $Q$-values of two possible actions for the same previous state: 
\begin{equation}
\small
    Q(\mathbf{s}_t, \mathbf{a}_t^{(1)}) \quad \text{\emph{v.s.}} \quad Q(\mathbf{s}_t, \mathbf{a}_t^{(2)})
\end{equation}
Next, we will introduce the detailed MCTS process to automatically derive the $Q$-values. Specifically, we will iterate through the following four operations until convergence.

\paragraph{Selection} Given the current tree $\mathcal{T}$, MCTS first needs to select a leaf node as a candidate for further exploration. 
By initializing the state $\mathbf{s}$ as the root, we use the PUCT criterion \citep{rosin2011multi} until a leaf node is encountered.
\begin{equation}
\small
    \arg\max_{\mathbf{a}} Q(\mathbf{s}, \mathbf{a}) + c_{\text{puct}} \frac{\sum_j \pi(a_j|\cdot)}{|\mathbf{a}|}\frac{\sqrt{N_{\text{parent}}(\mathbf{a})}}{1 + N(\mathbf{s}_t, \mathbf{a})}
\end{equation}
where $N(\cdot)$ represents the visit count, and $a_j$ is the $j$-th token in the step.

\paragraph{Expansion and Evaluation} Given the state represented by the selected leaf node, we sample multiple possible candidate actions for the next step. 
To encourage diversity, a higher temperature, typically ranging from 0.6 to 1, is used.

For efficient evaluation, we reuse the expanded nodes and simply apply a one-step rollout. 
If the rollout action is not terminal, we directly set the value of the current leaf node to 0. 
Otherwise, the final answer in the terminal action is evaluated for equivalence to the ground truth. 
If the final answer is correct, the reward $R$ will be 1; otherwise, it will be -1. 
Therefore, the value can be written as follows.
\begin{equation}
\small
V(\mathbf{s}) = \mathbb{I}_{\text{terminal}}(\mathbf{a}) \cdot R(\mathbf{s}, \mathbf{a})
\end{equation}
where $\mathbb{I}(\cdot)$ is the indicator function.

\paragraph{Backup} For the terminal nodes reached during the rollout and the current leaf node, MCTS performs a backward update of the visit count and $Q$-value for every $(\mathbf{s}^\prime, \mathbf{a}^\prime)$ along the path from current node to the root.
\begin{equation}
\small
   \begin{aligned}
    N(\mathbf{s}^\prime, \mathbf{a}^\prime) &\leftarrow N(\mathbf{s}^\prime, \mathbf{a}^\prime)  + 1 \\
    Q(\mathbf{s}^\prime, \mathbf{a}^\prime) &\leftarrow Q(\mathbf{s}^\prime, \mathbf{a}^\prime) + \frac{1}{N(\mathbf{s}^\prime, \mathbf{a}^\prime)} (V(\mathbf{s}) - Q(\mathbf{s}^\prime, \mathbf{a}^\prime))
\end{aligned} 
\end{equation}

As shown in Figure~\ref{fig:pipeline}, we obtain a solution tree $\mathcal{T}$ with many branches after running the above MCTS process for several iterations. 
From this tree, we can extract a partial solution and its two different next steps along with their corresponding $Q$-values. The step with the larger $Q$-value will be annotated as the preferred example.

\subsection{Step-level Preference Learning}\label{sec:step_pre_learn}

Given our autonomously generated step-level preference annotations, we propose an approach called step-level value preference optimization--\modelname. 
In contrast to DPO, we maintain 3 models with an additional value model $V_\phi$. 
Unlike in PPO, our value model is lightweight, achieved by adding an auxiliary value head directly over the policy model. 
This value head consists of a single linear layer with a \texttt{tanh} activation function, running parallel to the linear layer used for token prediction.

For notation simplification, we denote the annotated step-level preference instance $(\mathbf{s}_t, \mathbf{a}_t^w \succ \mathbf{a}_t^l)$ as $(\mathbf{s}_{t+1}^w \succ \mathbf{s}_{t+1}^l)$, where the two multi-step generations are only different at their last steps. 
According to our previous definition, the state $\mathbf{s}_{t+1}$ also represents the first $t$ steps $\mathbf{y}_{1:t}$~.

\paragraph{Pre-Training} In DPO, the policy model is pre-trained with a standard SFT loss. 
In our approach, due to the weights sharing architecture between $\pi$ and $V_{\phi}$, our pre-training adopts the multi-task loss.
\begin{equation}
\small
\begin{aligned}
    & \hat{V}(\mathbf{s}_{t+1}) = \begin{cases}
        R(\mathbf{s}_t, \mathbf{a}_t), & \text{$\mathbf{a}_t$ is terminal} \\
        Q(\mathbf{s}_t, \mathbf{a}_t), & \text{otherwise}
    \end{cases} \\
    &\mathcal{L} = \mathcal{L}_{\text{SFT}} + \mathbb{E}\left[ (V_\phi(\mathbf{s}_{t+1}) - \hat{V}(\mathbf{s}_{t+1}))^2 \right]
\end{aligned}
\label{eq:sft_mse}
\end{equation}
The mean squared error (MSE) loss is employed to pre-train the value head, which is also the pointwise approach in ranking algorithm~\citep{liu2009learning}. 
The label for the value prediction is either the $Q$-value of the intermediate step or the final reward.

\paragraph{\modelname} As indicated by DPO, the difference of implicit rewards for a pair of preference annotations can be re-parameterized as follows:
\begin{equation}
\small
    \Delta r_{\pi}(\mathbf{s}_{t+1}^w, \mathbf{s}_{t+1}^l) = \beta \log\frac{\pi (\mathbf{s}_{t+1}^w) \pi^\prime (\mathbf{s}_{t+1}^l)}{\pi^\prime (\mathbf{s}_{t+1}^w)\pi (\mathbf{s}_{t+1}^l)}
\label{eq:policy_diff}
\end{equation}
In our \modelname, we aim to optimize both policy and value models through preference learning. 
Accordingly, we define the explicit value difference.
\begin{equation}
\small
    \Delta r_{\phi}(\mathbf{s}_{t+1}^w, \mathbf{s}_{t+1}^l) = V_\phi(\mathbf{s}_{t+1}^w) - V_\phi(\mathbf{s}_{t+1}^l)
\label{eq:value_diff}
\end{equation}
We then propose the following \modelname loss function, which includes three different objectives.
\begin{equation}
\small
\begin{aligned}
&\mathcal{L}_{\modelname} =  -\log\sigma(\Delta r_{\pi}(\mathbf{s}_{t+1}^w, \mathbf{s}_{t+1}^l)) \\
& + \max(0, \gamma - \Delta r_{\phi}(\mathbf{s}_{t+1}^w, \mathbf{s}_{t+1}^l)) \\   
& + \left( \Delta r_{\pi}(\mathbf{s}_{t+1}^w, \mathbf{s}_{t+1}^l) -  \text{sg}\left[\Delta r_{\phi}(\mathbf{s}_{t+1}^w, \mathbf{s}_{t+1}^l)\right] \right)^2 
\end{aligned} 
\label{eq:svpo_loss}
\end{equation}
where the margin $\gamma \geq 0$ is tunable hyper-parameter, and $\text{sg}[\cdot]$ denotes the stop gradient operator. 

The first objective essentially replicates the original DPO loss $\mathcal{L}_{\text{DPO}}$ in (\ref{eq:dpo_loss}), applied to the automatically annotated step-level preference data. 

The second objective is a margin loss for value preference learning, inspired by the pairwise ranking algorithm~\citep{liu2009learning}. 
Given a non-negative margin $\gamma$, minimizing this loss encourages the value of the positive example to be larger than that of the negative one by at least $\gamma$. 
The detailed theoretical analysis can refer to \citep{chen2009ranking}.

The third objective is a regularization term adapted MSE loss, which aims to ensure a similar preference scale between the implicit reward model and our proposed explicit value model. 
In this loss, we use $\Delta r_{\phi}$ as the targeted label and detach its gradient to prevent model degeneration.

\paragraph{Analysis of Regularization} A natural question regarding the regularization term is whether designing the value output via \texttt{tanh} can match the reward defined in the log-likelihood ratio. 
We can first derive the possible theoretical matching range.
\begin{equation}
\small
\begin{aligned}
            & \Delta r_{\phi} \in [-2, 2] \\
\Rightarrow & \frac{\pi(\mathbf{y}^w)\pi^\prime(\mathbf{y}^l)}{\pi^\prime(\mathbf{y}^w)\pi(\mathbf{y}^l)} = e^{\Delta r_{\pi} / \beta} \in [e^{-2/\beta}, e^{2/\beta}]
\end{aligned}
\end{equation}
Therefore, the range is determined by $\beta$. 

\textbf{(1)} $\lim_{\beta\to0} [e^{-2/\beta}, e^{2/\beta}] = (0, +\infty)$: when $\beta$ is small in DPO or PPO, it can prevent the policy model from deviating too far. 
For example of the commonly used $\beta = 0.1$, the allowed matching range becomes $[e^{-20}, e^{20}]$, which is actually equivalent to $(0, +\infty)$ in the context of numerical precision. 
In other words, for smaller $\beta$, our regularization loss can easily match the scale between implicit and explicit preferences.

\textbf{(2)} $\lim_{\beta\to\infty} [e^{-2/\beta}, e^{2/\beta}] = \{1\}$: when $\beta$ becomes large, the allowed matching range will gradually center around 1, forcing the distance between $\pi$ and $\pi^\prime$ to be closer. 
In other words, for larger $\beta$, our regularization loss can also play the role of preventing the policy model from deviating too far.

\subsection{Step-level Inference}

Even without the value model, one can still directly apply greedy decoding to the policy model. 
However, incorporating the value model and an associated reranking criterion allows step-level beam search (SBS)~\citep{yu2023outcome,chen2024alphamath} to effectively select the preferred solution path in mathematical reasoning, all while incurring a lower computational cost compared to MCTS. 
Since our value preference learning is optimized at the step level and utilizes ranking loss, our approach seamlessly integrates with the inference framework of step-level beam search.

\section{Experiments}

\subsection{Experimental Setup}

We validate the applicability of our framework across various base models, including math domain-specific pre-trained models such as DeepseekMath-Base-7B~\citep{shao2024deepseekmath}, as well as general pre-trained models such as Llama3~\citep{llamma3blog}.
In this study, we mainly focus on how to improve mathematical reasoning skills through step-level preference learning.
Therefore, we obtain the corresponding multi-step SFT models using 27k MARIO seed data~\citep{liao2024mario} in XML format, as detailed in Appendix~\ref{sec:app_xml_format}.

\paragraph{Step-level Preference Annotation via MCTS} Given a multi-step SFT model for mathematical reasoning, we only extract the 15k questions from the GSM8K~\citep{abs-2110-14168} and MATH~\citep{HendrycksBKABTS21} datasets. 
Following the methodology described in Section~\ref{sec:mcts}, we employ the MCTS framework to automatically generate both multi-step solutions and step-level preferences. 
This process requires no supervision from either humans or GPT-4. 
Particularly, for each question, we continue constructing trees until we obtain four complete and correct multi-step solutions or until the number of trees reaches 10. 
We then extract step-level preferences from the trees in a top-down manner, maintaining an approximate ratio of 1:4 between positive and negative examples. 
Consequently, we acquire a total of 56k complete positive instances $\mathbf{y}^w$. 
Additional details are provided in Appendix~\ref{sec:app_data}.

\paragraph{Test sets} The in-domain test sets from GSM8K and MATH share the same distribution as our training data. 
Meanwhile, we evaluate our final checkpoint on the out-of-domain (OOD) datasets GaoKao2023~\citep{liao2024mario} and OCWCourses~\citep{LewkowyczADDMRS22}. 
These OOD test sets are even more challenging than the MATH dataset but inherently require multi-step reasoning. 

\paragraph{Baselines} For commercial and popular open-source models, we compared our approach with OpenAI's ChatGPT and GPT-4~\cite{openai2023gpt4}, Llama2~\cite{touvron2023llama}, and Llemma~\cite{azerbayev2023llemma} using the Chain of Thought (CoT)~\cite{Wei0SBIXCLZ22} and Program-Aided Language (PAL)~\cite{GaoMZ00YCN23}. 
Additionally, we benchmarked our method against recent high-performing fine-tuned mathematical LLMs, including MAmmoTH~\cite{yue2023mammoth}, MathCoder~\cite{wang2023mathcoder}, ToRA~\cite{gou2023tora}, MARIO~\cite{zhang2024mario}, MathGenie~\cite{lu2024mathgenie}, DeepSeekMath-Instruct~\cite{shao2024deepseekmath}, and AlphaMath~\citep{chen2024alphamath}. 
Similar to our approach, these models leverage a Python code interpreter for numerical calculations. 
Further implementation details are provided in Appendix~\ref{sec:app_implement}.

\begin{table*}[!ht]
\centering
\small
    \resizebox{0.95\linewidth}{!}{
\begin{tabular}{@{}lccccccc@{}}
\toprule
\multirow{2}{*}{\textbf{Model}} & \multirow{2}{*}{\textbf{Size}} & \multirow{2}{*}{\textbf{Tool}} & \multicolumn{1}{c|}{\multirow{2}{*}{\textbf{\begin{tabular}[c]{@{}c@{}}Zero\\ Shot\end{tabular}}}} & \multicolumn{2}{c|}{\textbf{In-Domain}}             & \multicolumn{2}{c}{\textbf{OOD}} \\
                                &                                &                                & \multicolumn{1}{c|}{}                                                                              & \textbf{GSM8k} & \multicolumn{1}{c|}{\textbf{MATH}} & \textbf{GK2023}  & \textbf{OCW}  \\ \midrule
\multicolumn{8}{c}{Proprietary Models}                                                                                                                                                                                                                                                          \\ \midrule
GPT-4                           & -                              & \xmark                         & \multicolumn{1}{c|}{\xmark}                                                                        & 92.0           & \multicolumn{1}{c|}{42.5}          & -                & -             \\
GPT-4 (PAL)                     & -                              & \cmark                         & \multicolumn{1}{c|}{\xmark}                                                                        & 94.2           & \multicolumn{1}{c|}{69.7}          & 43.6             & 30.1          \\
ChatGPT                         & -                              & \xmark                         & \multicolumn{1}{c|}{\xmark}                                                                        & 80.8           & \multicolumn{1}{c|}{35.5}          & -                & -             \\
ChatGPT (PAL)                   & -                              & \cmark                         & \multicolumn{1}{c|}{\xmark}                                                                        & 78.6           & \multicolumn{1}{c|}{38.7}          & -                & -             \\
\midrule
\multicolumn{8}{c}{Open-Source Models}                                                                                                                                                                                                                                                          \\ \midrule
Llama-2                         & 7B                             & \xmark                         & \multicolumn{1}{c|}{\xmark}                                                                        & 13.3           & \multicolumn{1}{c|}{4.1}           & -                & 3.7           \\
CodeLlama                       & 7B                             & \xmark                         & \multicolumn{1}{c|}{\xmark}                                                                        & 10.5           & \multicolumn{1}{c|}{4.5}           & -                & 4.7           \\
CodeLlama(PAL)                  & 7B                             & \cmark                         & \multicolumn{1}{c|}{\xmark}                                                                        & 27.1           & \multicolumn{1}{c|}{17.2}          & -                & -             \\
Llemma                          & 7B                             & \xmark                         & \multicolumn{1}{c|}{\xmark}                                                                        & 36.4           & \multicolumn{1}{c|}{18.0}          & -                & 7.7           \\
Llemma (PAL)                    & 7B                             & \cmark                         & \multicolumn{1}{c|}{\xmark}                                                                        & 40.1           & \multicolumn{1}{c|}{21.5}          & -                & -             \\
DeepSeekMath-Base(PAL)          & 7B                             & \cmark                         & \multicolumn{1}{c|}{\xmark}                                                                        & 66.9           & \multicolumn{1}{c|}{31.4}          & -                & -             \\ \midrule
\multicolumn{8}{c}{Tuning Models}                                                                                                                                                                                                                                                               \\ \midrule
MAmmoTH-Coder                   & 34B                            & \cmark                         & \multicolumn{1}{c|}{\cmark}                                                                        & 72.7           & \multicolumn{1}{c|}{43.6}          & 25.2             & 14.0          \\
MathCoder                       & 34B                            & \cmark                         & \multicolumn{1}{c|}{\cmark}                                                                        & 81.7           & \multicolumn{1}{c|}{46.1}          & -                & -             \\
ToRA-Code                       & 34B                            & \cmark                         & \multicolumn{1}{c|}{\cmark}                                                                        & 80.7           & \multicolumn{1}{c|}{50.8}          & 31.7             & 5.5           \\
MARIO                           & 34B                            & \cmark                         & \multicolumn{1}{c|}{\cmark}                                                                        & 78.2           & \multicolumn{1}{c|}{53.5}          & 42.6             & 30.2          \\
MathGenie                       & 34B                            & \cmark                         & \multicolumn{1}{c|}{\cmark}                                                                        & 84.1           & \multicolumn{1}{c|}{55.1}          & -                & -             \\ \midrule
Llama-2 SFT                     & 7B                             & \xmark                         & \multicolumn{1}{c|}{\cmark}                                                                        & 41.3           & \multicolumn{1}{c|}{7.2}           & -                & -             \\
Llama-2 RFT                     & 7B                             & \xmark                         & \multicolumn{1}{c|}{\cmark}                                                                        & 51.2           & \multicolumn{1}{c|}{-}             & -                & -             \\
MAmmoTH-Coder                   & 7B                             & \cmark                         & \multicolumn{1}{c|}{\cmark}                                                                        & 59.4           & \multicolumn{1}{c|}{33.4}          & 15.3             & 11.0          \\
MathCoder                       & 7B                             & \cmark                         & \multicolumn{1}{c|}{\cmark}                                                                        & 67.8           & \multicolumn{1}{c|}{30.7}          & -                & -             \\
ToRA                            & 7B                             & \cmark                         & \multicolumn{1}{c|}{\cmark}                                                                        & 68.8           & \multicolumn{1}{c|}{40.1}          & 19.5             & 2.6           \\
ToRA-Code                       & 7B                             & \cmark                         & \multicolumn{1}{c|}{\cmark}                                                                        & 72.6           & \multicolumn{1}{c|}{44.6}          & 23.9             & 4.8           \\
MARIO                           & 7B                             & \cmark                         & \multicolumn{1}{c|}{\cmark}                                                                        & 74.5           & \multicolumn{1}{c|}{48.3}          & 34.5             & 21.7          \\
MathGenie                       & 7B                             & \cmark                         & \multicolumn{1}{c|}{\cmark}                                                                        & 76.0           & \multicolumn{1}{c|}{48.3}          & -                & -             \\
DeepSeekMath-Instruct           & 7B                             & \cmark                         & \multicolumn{1}{c|}{\cmark}                                                                        & \textbf{83.7}           & \multicolumn{1}{c|}{57.4}          & 43.9             & 18.0          \\
AlphaMath                       & 7B                             & \cmark                         & \multicolumn{1}{c|}{\cmark}                                                                        & 73.5           & \multicolumn{1}{c|}{53.6}          & 40.5             & 26.1          \\
\multicolumn{2}{l}{\quad + SBS ($B_1 = 1$)}                      & \cmark                         & \multicolumn{1}{c|}{\cmark}                                                                        & 81.1           & \multicolumn{1}{c|}{62.8}          & 46.2             & 30.5          \\
\multicolumn{2}{l}{\quad + SBS ($B_1 = 3$)}                      & \cmark                         & \multicolumn{1}{c|}{\cmark}                                                                        & 84.1           & \multicolumn{1}{c|}{66.3}          & 51.4             & 33.1          \\ \midrule
\modelname (Ours)               & 7B                             & \cmark                         & \multicolumn{1}{c|}{\cmark}                                                                        & 81.7           & \multicolumn{1}{c|}{\textbf{59.5}}          & \textbf{47.1}             &    \textbf{34.2}           \\
\multicolumn{2}{l}{\quad + SBS ($B_1 = 1$)}                      & \cmark                         & \multicolumn{1}{c|}{\cmark}                                                                        & 85.9           & \multicolumn{1}{c|}{64.4}          & 54.6             &    36.8           \\
\multicolumn{2}{l}{\quad + SBS ($B_1 = 3$)}                      & \cmark                         & \multicolumn{1}{c|}{\cmark}                                                                        & \cellcolor[HTML]{DAE8FC}\textbf{86.5}  & \multicolumn{1}{c|}{\cellcolor[HTML]{DAE8FC}\textbf{67.2}} & \cellcolor[HTML]{DAE8FC}\textbf{55.3}    &         \cellcolor[HTML]{DAE8FC}\textbf{40.8}      \\ \bottomrule
\end{tabular}
}
\caption{Main results. The best results for greedy decoding and step-level beam search (SBS) are highlighted in bold and \colorbox[HTML]{DAE8FC}{blue box}, respectively. By default, we set the beam size $B_2=5$ in SBS.}
  \label{tab:main_results}
\vspace{-1em}
\end{table*}

\subsection{Main Results}
For a fair comparison, in Table~\ref{tab:main_results}, we report the in-domain and out-of-domain results of our \modelname based on DeepSeekMath-Base-7B, which is consistent with the state-of-the-art methods, such as DeepSeekMath-Instruct~\citep{shao2024deepseekmath} and AlphaMath~\citep{chen2024alphamath}.

\paragraph{Greedy Decoding} Without the assistance of a value model, we first evaluate the policy model using greedy decoding, which is comparable to most related works. 
The main conclusion is that for more difficult problems requiring more reasoning steps, our approach shows greater advantages. 
As the difficulty increases for GSM8K, MATH, GaoKao2023 (GK2023), and OCWCourses (OCW), our approach achieves improvements of -2.0\% / +2.1\% / +3.2\% / +16.2\% over the previous state-of-the-art, DeepSeekMath-Instruct.

We slightly lag behind in GSM8K, which could be attributed to two possible reasons. 
First, GSM8K usually requires single step solution and less logical reasoning.
Second, the diversity of our training dataset is limited. 
While DeepSeekMath utilized 776k high-quality supervised data, we only autonomously generated 56k complete positive examples based on 15k questions.

\paragraph{SBS} With the value model optimized by step-level value preference learning, we can utilize the computationally efficient step-level beam search (SBS) to investigate the role of the value model in facilitating mathematical reasoning. 
Compared to greedy decoding, the value model significantly assists the policy model in navigating more effective reasoning paths, rather than solely relying on prior probabilities. 
Compared with AlphaMath, our \modelname achieves an average improvement of 5.3\% / 3.7\% on $B_1=1$ / $B_1=3$, respectively, which demonstrates the effectiveness of our approach.
We will further analyze the value model in subsequent ablation studies.
It is worth noting that with the help of the value model, our \modelname on 7B LLMs achieves comparable or even better results than GPT-4 in the challenging datasets.

\subsection{Analysis 1: Policy Model}
In this section, we will investigate the impact of step-level preferences on the policy model and explore the performance of different base models in our \modelname framework.

\begin{table}[t]
\centering
    \resizebox{\linewidth}{!}{
\begin{tabular}{@{}lcccc@{}}
\toprule
\multirow{2}{*}{\textbf{\begin{tabular}[c]{@{}l@{}}Training\\ Paradigm\end{tabular}}} & \multicolumn{2}{c}{\textbf{In-Domain}} & \multicolumn{2}{c}{\textbf{OOD}} \\
                                                                                      & \textbf{GSM8k}     & \textbf{MATH}     & \textbf{GK2023}  & \textbf{OCW}  \\ \midrule
SFT                                                                                   & 77.7               & 56.9              & 43.1             & 27.5             \\
DPO$^\dag$                                                                                   & 78.9               & 57.1              & 45.4             &       28.3        \\ \midrule
\modelname (Ours)                                                                            & 81.7               & 59.5              & 47.1             &      34.2         \\
- w/o regularization                                                                     & 80.2               & 58.3              & 46.2             &    32.1           \\ \bottomrule
\end{tabular}
}
\caption{Ablation study of training paradigm on policy model. $^\dag$Solution-level DPO.}
  \label{tab:ablation_policy}
\end{table}

\begin{table}[]
\centering
\setlength{\tabcolsep}{2pt}
\resizebox{\linewidth}{!}{
\begin{tabular}{@{}lcccc@{}}
\toprule
\multirow{2}{*}{\textbf{Model}} & \multicolumn{2}{c}{\textbf{In-Domain}} & \multicolumn{2}{c}{\textbf{OOD}} \\
                                & \textbf{GSM8K}     & \textbf{MATH}     & \textbf{GK2023}  & \textbf{OCW}  \\ \midrule
Llama3-8B + SFT                    & 75.9               & 46.5              & 33.2             & 10.3          \\
Llama3-8B-Instruct              & 79.6               & 30.0              & -                & -             \\
Llama3-70B-Instruct             & \textbf{93.0}      & 50.4              & -                & -             \\ \midrule
Llama3-8B + \modelname          & 81.3               & 48.8              & 35.6             & 11.1          \\
\quad + SBS ($B_1=1$)           & 84.3               & 54.2              & 40.0             & 13.3          \\
\quad + SBS ($B_1=3$)           & 85.5               & \textbf{56.3}     & \textbf{43.7}    & \textbf{16.6} \\ \bottomrule
\end{tabular}
}
\caption{Performance comparison of Llama3 series.}
  \label{tab:llama3}
\end{table}

\paragraph{Ablation Study of Training Paradigm} As shown in Table~\ref{tab:ablation_policy}, we compare the performance of the policy model under different preference optimization. 
Our principal findings are as follows:
\textbf{(1)} Compared to SFT, which blindly imitates positive examples $\mathbf{y}^w$, preference learning encourages the policy model to distinguish between $\mathbf{y}^w$ and $\mathbf{y}^l$, thereby enhancing its reasoning capability. 
However, solution-level DPO is limited by its coarse preference relationships, which do not indicate which specific step in the negative solutions $\mathbf{y}^l$ led to the mistakes.
\textbf{(2)} Compared to solution-level DPO, our proposed step-level preferences can significantly enhance the reasoning performance of the policy model on both in-domain and out-of-domain datasets.
This can be attributed to the more granular information of reasoning steps reflected by $Q$-value in the Monte Carlo tree.
\textbf{(3)} The value model can further guide the optimization of the policy model, as evidenced by the performance decreases when the regularization term is removed.

\paragraph{Discussion of Different Base Models} We further investigate the performance of the general pre-trained model, Llama3~\citep{llamma3blog}, within our framework.
As shown in Table~\ref{tab:llama3}, we have the following main findings:
\textbf{(1)} Compared to DeepSeekMath-Base-7B in Table~\ref{tab:main_results}, the overall performance of the general pre-trained model Llama3 is relatively insufficient.
This is because DeepSeekMath-Base is pre-trained on a substantial math-related corpus and is believed to process more necessary mathematical knowledge, resulting in higher quality preference data.
\textbf{(2)} Our \modelname outperforms the instruction-tuned Llama3 and approached the performance of the 70B model.
Furthermore, compared to the SFT model, we achieved significant improvements, demonstrating the effectiveness and applicability of our approach.

\subsection{Analysis 2: Value Model}\label{sec:analysis_value_model}
In this section, we further investigate the impact of value loss (mainly including the MSE loss in Eq.~(\ref{eq:sft_mse}) and Margin loss in Eq.~(\ref{eq:svpo_loss}) on performance and evaluate the accuracy of identifying preferences.

\begin{table}[t]
\centering
\setlength{\tabcolsep}{2pt}
    \resizebox{\linewidth}{!}{
\begin{tabular}{@{}lccccc@{}}
\toprule
\multirow{2}{*}{\textbf{Method}}   & \multirow{2}{*}{\textbf{SBS}} & \multicolumn{2}{c}{\textbf{In-Domain}} & \multicolumn{2}{c}{\textbf{OOD}} \\
                                   &                               & \textbf{GSM8K}     & \textbf{MATH}     & \textbf{GK2023}  & \textbf{OCW}  \\ \midrule
\multirow{2}{*}{\modelname (Ours)} & $B_1 = 1$                     & 85.9               & 64.4              & 54.6             &       36.8        \\
                                   & $B_1 = 3$                     & 86.5               & 67.2              & 55.3             &       40.8        \\ \midrule
\multirow{2}{*}{w/o Margin loss}   & $B_1 = 1$                     & 85.4               & 62.5              & 49.6             &      34.9         \\
                                   & $B_1 = 3$                     & 85.2               & 63.7              & 52.2             &       37.5        \\ \midrule
\multirow{2}{*}{w/o MSE loss}      & $B_1 = 1$                     & 83.8               & 60.5              & 52.2             &        30.8       \\
                                   & $B_1 = 3$                     & 82.1               & 56.7              & 45.7             &       28.6        \\ \bottomrule
\end{tabular}
}
\caption{Ablation study of value model.}
  \label{tab:ablation_value}
\end{table}

\paragraph{Ablation Study of Value Loss} As shown in Table~\ref{tab:ablation_value}, we compare the performance of step-level beam search in different setups for value loss.
Our principal findings are as follows:
\textbf{(1)} When margin loss is omitted, which describes the local relationships in preference data, the performance will decrease. 
As explained in the method, this can be attributed to the ability of margin loss to further distinguish the value of candidate actions. 
\textbf{(2)} MSE loss is crucial for the value model, as it provides global information for each node in the Monte Carlo tree.
Relying solely on preference relationships by margin loss may cause the value model to lose the ability to screen cousin nodes (\emph{i.e.}, candidate actions at the same level but with different previous states). 
This explains why the performance of $B_1 = 3$ is significantly lower than that of $B_1 = 1$ when MSE loss is omitted.
In summary, MSE loss and margin loss provide complementary information, and their combined effect leads to a better value model.

\begin{figure}[b]
\centering
    \begin{minipage}[c]{\linewidth}
        \centering
        \begin{subfigure}[b]{0.495\linewidth}
            \centering
            \includegraphics[width=\linewidth]{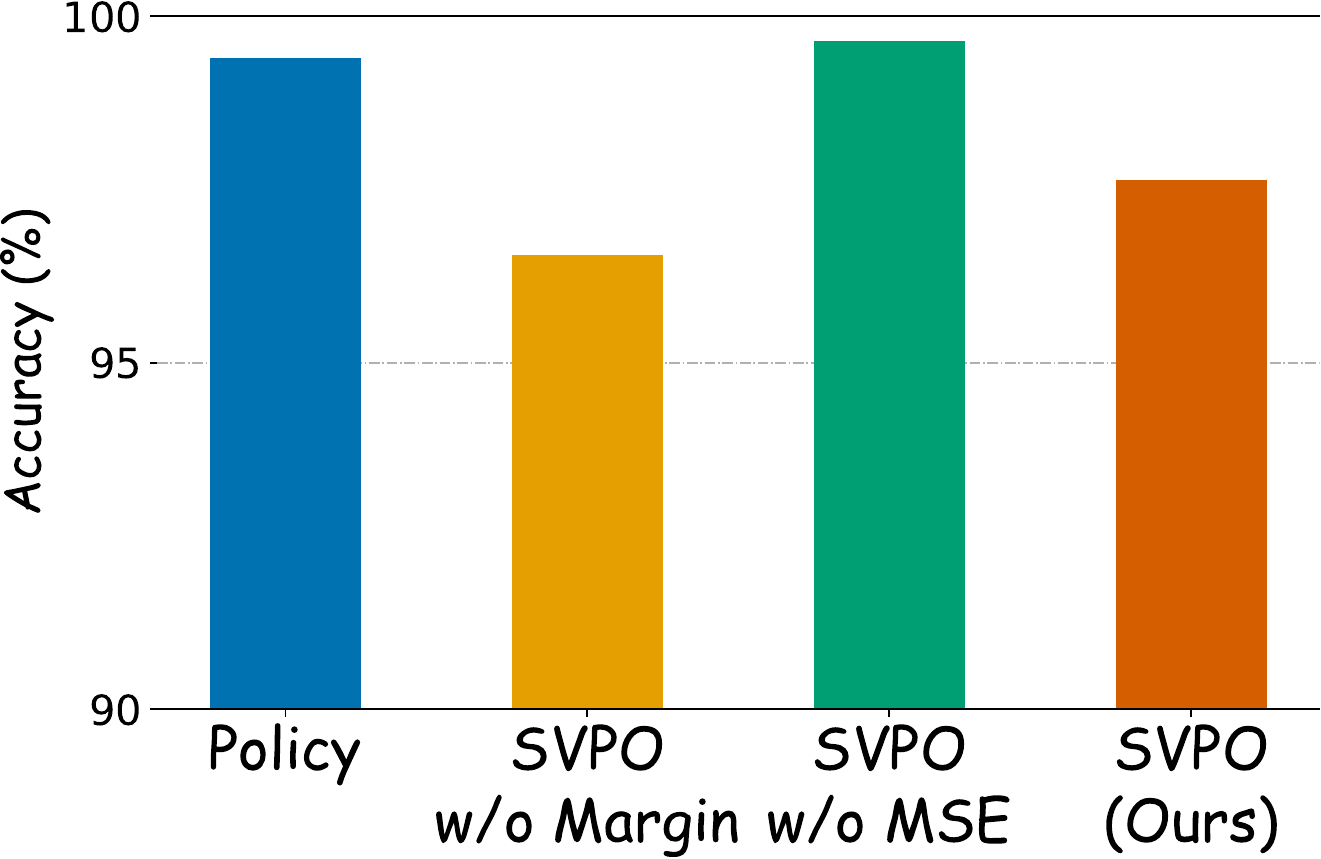}
            \caption{Training Set}
            \label{fig:win_rate_a}
        \end{subfigure}
        \hspace{-0.2cm}
        \begin{subfigure}[b]{0.49\linewidth}
            \centering
            \includegraphics[width=\linewidth]{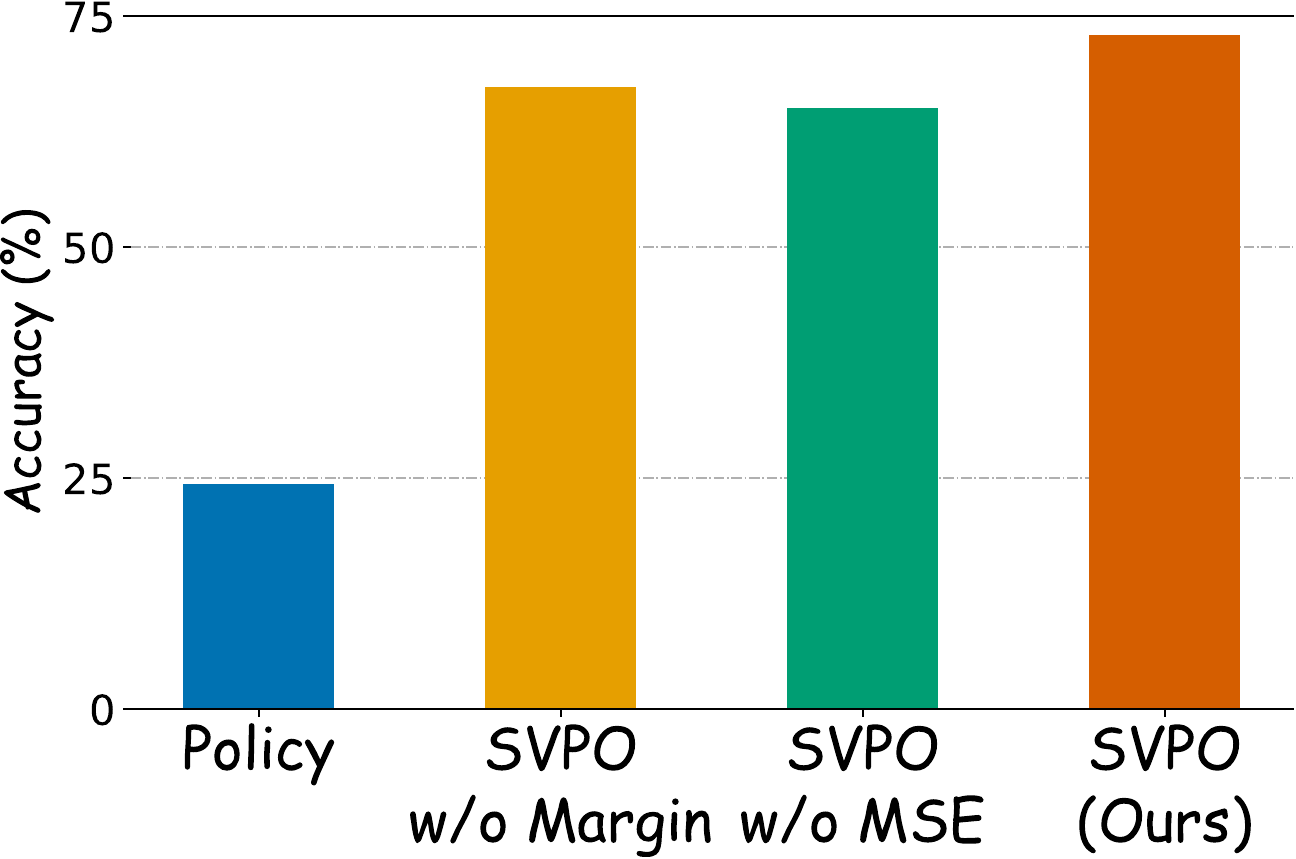}
            \caption{Test set}
            \label{fig:win_rate_b}
        \end{subfigure}
        \caption{Win Rate of Preference.}
        \label{fig:win_rate}
    \end{minipage}%
\end{figure}

\paragraph{Win Rate of Preference} We conduct a further investigation into the accuracy of the policy and value model by Eq.~(\ref{eq:policy_diff}) and~(\ref{eq:value_diff}) in assessing preference relationships.
We randomly select 200 questions from the test sets of GSM8K and MATH respectively, and utilize MCTS to build preferences as the test set in the win rate.
As shown in Figure~\ref{fig:win_rate}, we have the following main findings:
\textbf{(1)} The preference relationships in training sets can be easily mastered, as evidenced in Figure~\ref{fig:win_rate_a}, where the accuracy of both ``Policy'' and ``\modelname w/o MSE'' significantly surpasses that of others.
However, the poor performance of ``Policy'' on the test set indicates that the implicit reward model (\emph{i.e.}, policy model) is highly susceptible to overfitting.
\textbf{(2)} Compared to the implicit reward model, our proposed explicit value model is relatively stable even if it only learns preference relationships by margin loss.
This further demonstrates the effectiveness of our value model.

\subsection{Sensitivity of $\beta$ and $\gamma$}

\begin{figure}[h]
\centering
    \begin{minipage}[c]{\linewidth}
        \centering
        \begin{subfigure}[b]{0.50\linewidth}
            \centering
            \includegraphics[width=\linewidth]{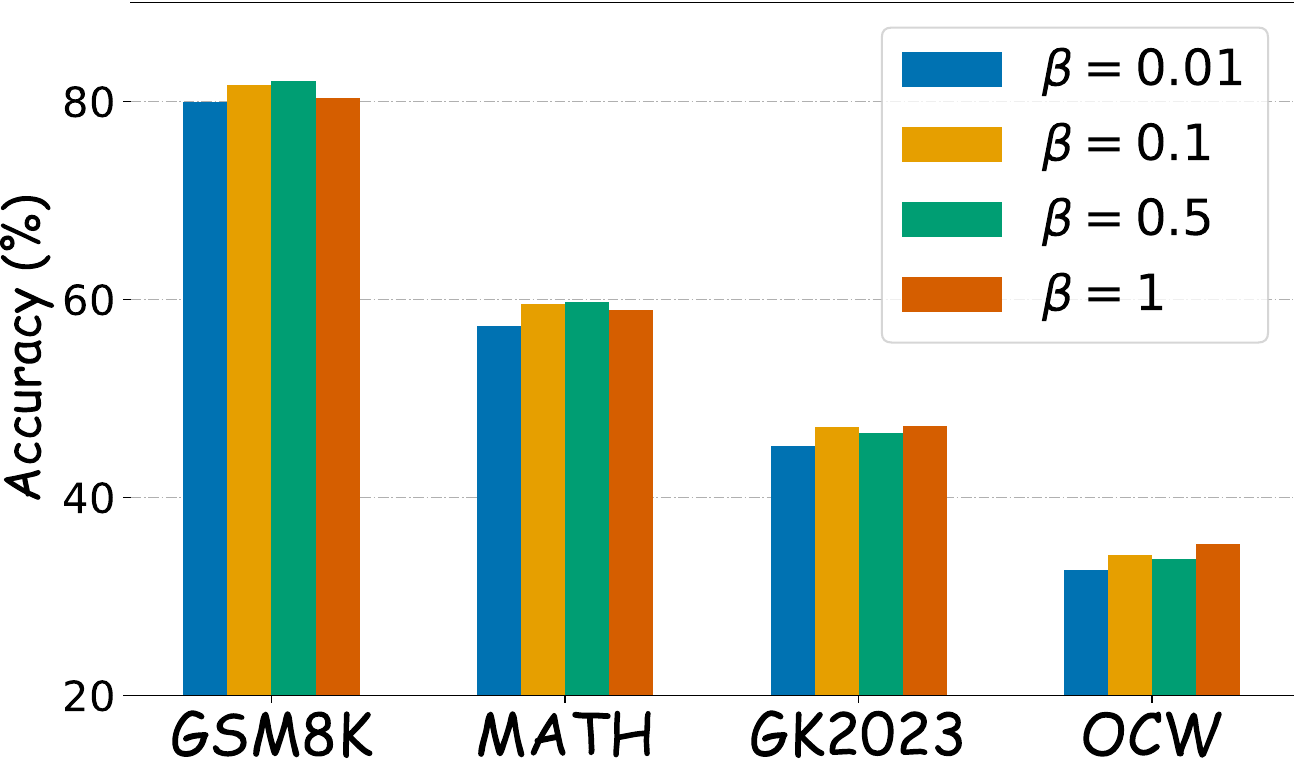}
            \caption{Beta $\beta$}
            \label{fig:sensitive_beta}
        \end{subfigure}
        \hspace{-0.2cm}
        \begin{subfigure}[b]{0.49\linewidth}
            \centering
            \includegraphics[width=\linewidth]{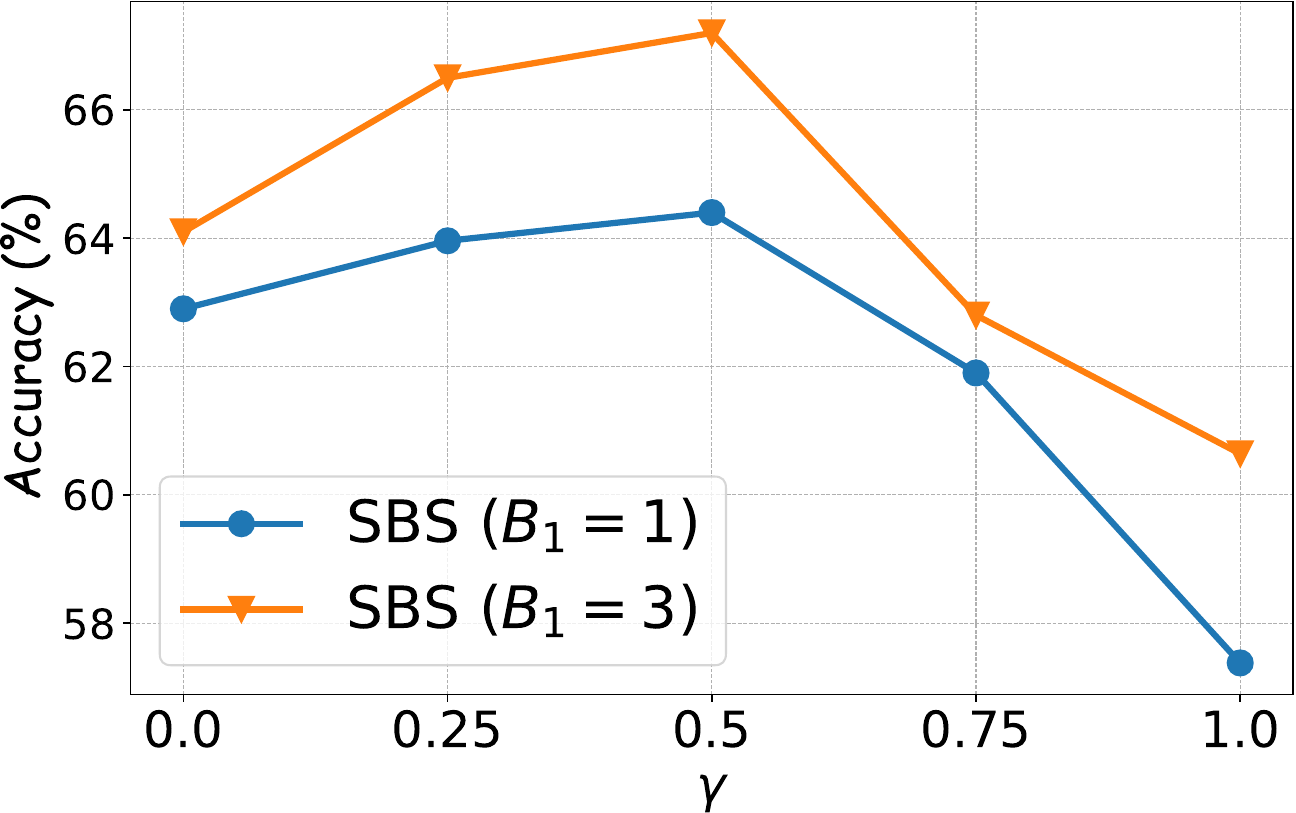}
            \caption{Margin $\gamma$}
            \label{fig:sensitive_gamma}
        \end{subfigure}
        \caption{Hyperparameter sensitivity analysis.}
        \label{fig:sensitive_analysis}
    \end{minipage}%
\end{figure}

$\beta$ in Eq.~(\ref{eq:policy_diff}) controls the implicit reward model, while the margin $\gamma$ in Eq.~(\ref{eq:svpo_loss}) controls the explicit value model. 
Thus, we investigate the impact of the two key hyper-parameters. 

\paragraph{Beta $\beta$} Following DPO~\citep{RafailovSMMEF23}, we investigate the impact of different $\beta$ on the policy model, as shown in Figure~\ref{fig:sensitive_beta}.
We observe that the optimization of the policy model remains relatively stable across different $\beta$.
This can be attributed to the regularization term, as analyzed in Section~\ref{sec:step_pre_learn}.
The explicit value model can prevent the policy model $\pi$ from deviating too far from the reference policy model $\pi^\prime$ through the regularization term, thereby improving the stability.

\paragraph{Margin $\gamma$}
As shown in Figure~\ref{fig:sensitive_gamma}, we evaluate the results of SBS in MATH with varying $\gamma$ between $[0,1]$.
We observe that an excessively large $\gamma$ will cause the value model to degenerate. This can be attributed to the fact that a large margin $\gamma$ compresses the predicted values towards either -1 or 1, making it difficult for the value model to correctly differentiate between candidate actions at the same level in SBS.
Moreover, setting $\gamma$ to 0 also leads to a performance degradation, indicating that it is not the optimal target margin. Although with $\gamma = 0$ the value model still maintains the value preference learning, an appropriate gamma is conducive to increase the confidence in scoring and enhances the model's generalization.

\section{Related Work}

\paragraph{Mathematical Reasoning}
Recent work~\citep{gou2023tora,liao2024mario,lu2024mathgenie,shao2024deepseekmath} has achieved remarkable progress in mathematical reasoning.
However, most efforts focus solely on supervised fine-tuning, which makes LLMs blindly imitate positive solutions without understanding what the wrong solutions are.

\paragraph{Preference Learning}
Recently, preference learning~\citep{RafailovSMMEF23,abs-2402-01306,abs-2402-05369} has attracted significant attention due to its ability to align with human preferences and distinguish between positive and negative examples.
However, due to focusing solely on coarse solution-level preferences, most existing work is limited in performance on multi-step reasoning tasks, particularly in mathematical reasoning.
Compared to previous work, our \modelname autonomously annotates step-level preferences through MCTS, and reflects potential reasoning errors through the $Q$-values at each step, thereby significantly improving the performance of preference learning on multi-step reasoning tasks.

\paragraph{Value Model}
The value model is derived from the state-value function in reinforcement learning (RL), which is used to evaluate the expected return of the current state.
Recent work~\citep{abs-2309-15028} has found that the value model can effectively enhance the reasoning capability of the policy model but limited by the complex training process of RL.
In our study, we propose step-level value preference optimization, which achieves higher quality value models in a simpler training process.

\section{Conclusion}

In this study, we introduce \textit{Step-level Value Preference Optimization} (\modelname) by extending Direct Preference Optimization (DPO) through the integration of a lightweight step-level value model. 
The training framework of \modelname is much more computationally efficient compared to Proximal Policy Optimization (PPO). 
Extensive experimental results demonstrate that for tasks involving multi-step mathematical reasoning, our approach significantly enhances performance, particularly with the support of the proposed value model.


\section*{Limitations}

First, we consider our work, \modelname, as a trade-off approach between DPO and PPO, offering a relatively lower computational cost. 
Beyond being an Empirical Method in Natural Language Processing (EMNLP), the theoretical foundation of margin loss in the area of learning-to-rank has been extensively discussed in~\citet{chen2009ranking}, and we also theoretically analyze how the regularization loss in $\mathcal{L}_\modelname$ impacts the preference learning of the policy model. 
Nevertheless, in the future work, we need to establish a more solid theoretical foundation to connect the implicit reward model and the explicit value model.

Secondly, although our method has achieved excellent performance in multi-step reasoning, particularly in mathematical reasoning, there is still an issue that deserves further exploration in future work: whether our \modelname can enhance mathematical reasoning capabilities in the context of multimodal data.
This may include mathematical reasoning from multiple combinations of modalities, such as language, images, tables, or audio, which is an increasingly prevalent and demanding type of reasoning in real-world scenarios.
In future work, we plan to extend our \modelname to accommodate multimodal scenarios.

Additionally, although in this study we integrate the step-level value preference optimization into DPO as an example, our approach is broadly applicable to various types of preference learning algorithms~\citep{abs-2402-01306,abs-2402-05369,abs-2403-07691}. 
In future work, we will explore incorporating our \modelname into these preference learning algorithms.

\section*{Ethics Statement}
This work primarily focuses on mathematical reasoning tasks, and our contributions are entirely methodological. Therefore, this work does not have direct negative social impacts.
For the experiments, we have open-sourced the code and utilized openly available datasets commonly used in previous research, without any sensitive information to our knowledge.
The authors of this work adhere to the ACL ethical guidelines, and the application of this work does not present any apparent issues that may lead to ethical risks.

\section*{Acknowledgments}
This work was supported by Alibaba Research Intern Program.


\bibliography{custom}

\newpage
\appendix

\section{Implementation Details}
\label{sec:app_implement}

\subsection{Detailed Setup}

\textbf{For step-level preference annotation via MCTS}, we set $c_{\text{puct}}$ to 1.25, set the temperature within the range of 0.6 to 1, limit the maximum tree depth to 8, set each node to expand 5 child nodes, and simulate at most 60 times.
For each question in training set, we construct at most 10 trees.
Following~\citet{chen2024alphamath}, we define two types of steps in MCTS, $\mathcal{C}$-step and $\mathcal{A}$-step.
The $\mathcal{C}$-step is responsible for code execution and consists of text analysis, code snippets, and execution results.
The $\mathcal{A}$-step is responsible for summarizing the answers, comprising text analysis and the final answer.
We organize these two steps in the following XML format:

\begin{center}
\begin{tcolorbox}[title=$\mathcal{C}$-step, width=\columnwidth]
\begin{small}
\begin{Verbatim}[commandchars=\\\{\}]
<step>
<p>
\texttt{\textcolor{red}{\{textual analysis\}}}
</p>
<code>
\texttt{\textcolor{red}{\{code snippets\}}}
</code>
<p>
\texttt{\textcolor{red}{\{code output\}}}
</p>
</step>
\end{Verbatim}
\end{small}
\end{tcolorbox}
\end{center}

\begin{center}
    \begin{tcolorbox}[title=$\mathcal{A}$-step, width=\columnwidth]

\begin{small}
\begin{Verbatim}[commandchars=\\\{\}]
<step>
<p>
\texttt{\textcolor{red}{\{textual analysis\}}}
</p>
<p>
\texttt{Final Answer: \textcolor{red}{\{predicted answer\}}}
</p>
</step>
\end{Verbatim}
\end{small}
\end{tcolorbox}
\end{center}

\textbf{For Pre-training via SFT}, we convert the pre-trained model into a corresponding multi-step SFT model through the pre-training loss in Eq.~(\ref{eq:sft_mse}).
We set the learning rate to 2e-5, the batch size to 512, fix the MSE weight to 0.01, and train for 10 epochs.
We employ the AdamW optimizer~\citep{LoshchilovH19} and a cosine learning rate scheduler, setting the warm-up rate to 0.03.

\textbf{For \modelname}, we set $\beta$ to 0.1, $\gamma$ to 0.5, learning rate to 5e-6, batch size to 512, and train for 1 epoch.
Since preference learning may easily degenerate model, it is common practice to incorporate SFT loss in RLHF or DPO training~\citep{Ouyang0JAWMZASR22,pang2024iterative} to mitigate this issue.
Thus, we also use the pre-training loss including standard SFT loss and MSE value loss in preference optimization stage.
Specifically, we fixed the weights for the margin loss and MSE loss at 0.25, the weight for the regularization term at 0.001, and the weight for the SFT loss at 5.
In addition, we also employ the AdamW optimizer~\citep{LoshchilovH19} and the cosine learning rate scheduler with a warmup rate of 0.03.

\subsection{Datasets Details}
\label{sec:app_data}

\begin{table}[t]
\centering
\setlength{\tabcolsep}{2pt}
    \resizebox{0.95\linewidth}{!}{
  \centering
\begin{tabular}{@{}lccc@{}}
\toprule
Dataset                                 & OOD?      & \# Training & \# Test \\ \midrule
GSM8K            & In-Domain & 7473       & 1319   \\
MATH          & In-Domain & 7500       & 5000   \\
GaoKao2023       & OOD       & -          & 385    \\
OCWCourses & OOD       & -          & 272    \\ \bottomrule
\end{tabular}
}
\caption{Datasets Statistics}
\label{tab:appendix_dataset_statistic}
\end{table}

\paragraph{Mathematical Reasoning Benchmarks}
Table~\ref{tab:appendix_dataset_statistic} provides a detailed overview of the mathematical reasoning benchmarks. The training and test sets are divided in accordance with previous studies~\citep{abs-2110-14168,HendrycksBKABTS21}. GSM8K~\citep{abs-2110-14168} is a dataset focused on multi-step mathematical reasoning, featuring high-quality, diverse grade school math word problems crafted by human authors. The MATH dataset~\citep{HendrycksBKABTS21} contains complex competitive mathematics problems. GaoKao2023~\citep{liao2024mario} includes math problems from the 2023 Chinese National College Entrance Examination, the 2023 American Mathematics Competitions, and the 2023 American College Testing. OCWCourses~\citep{LewkowyczADDMRS22} is a compilation of 272 STEM problems targeted at the undergraduate level, most of which require multi-step reasoning.

\paragraph{Preference Test set in the Win Rate}
In Section~\ref{sec:analysis_value_model}, we evaluate the accuracy of the policy model and the value model in assessing preferences.
These models can accurately assess the preferences on the training set, as shown in Figure~\ref{fig:win_rate_a}.
To accurately evaluate the generalization of the value model, we randomly sample 200 questions from the test sets of GSM8K and MATH respectively, and constructed total 10633 preference pairs using MCTS.

\paragraph{Step-level Preference Pairs Construction in Monte Carlo tree}
After step-level preferences annotation via MCTS, we need to filter the preference pairs from the Monte Carlo tree for training.
Given step-level beam search, we need to consider the preference relationships among sibling nodes (at the same layer with same previous state), cousin nodes (at the same layer but with different previous states), and non-same-level terminal nodes (at different layer with terminal nodes).

Algorithm~\ref{alg:construct_pairs} outlines the process of our step-level preference pairs construction.
First, we label each node along the path based on the correctness of the terminal node (Lines 2-4).
Then, we iteratively construct step-level preference pairs in a top-down manner (Lines 5-23).
In this process, we can specify the quantities of the three different types of preference relationships.
In this study, we set the number of sibling nodes to 2, the number of cousin nodes and non-same-level terminal nodes to 1, respectively.
This maintains an approximate ratio of 1:4 between positive and negative examples.

\begin{algorithm*}[htbp]
\caption{Step-level Preference Pairs Construction}
\label{alg:construct_pairs}
\begin{algorithmic}[1]
\Require Monte Carlo trees $\mathcal{T}$, prompted question $\mathbf{x}$.
\Ensure Step-level Preference Pairs $\mathcal{P}$.
\State $\mathcal{P} = [~]$   \Comment{Initialization}
\For{terminal node $n$ in $\mathcal{T}$}
    \If{$n$ has correct final answer}
        \State Backpropagation labels each node as \textit{correct} along the path from root to $n$
    \EndIf
\EndFor
\State $n \leftarrow$ root node in $\mathcal{T}$
\While{$n$ is non-terminal node} \Comment{In a top-down manner}
\State $n_{w} \leftarrow \arg\max_{child \in n.children} \{ Q(child) | child \text{ is correct node.}\}$ \Comment{Ensure $\mathbf{y}^w$ is a correct solution}
\State $\mathbf{y}^w \leftarrow$ partial solution from root to $n_{w}$ in $\mathcal{T}$ 
\State $n_s \leftarrow$ randomly select a non-correct node in $n.children$  \Comment{Pairs in sibling nodes}
\If{$n_s \neq \emptyset$}
    \State $\mathbf{y}^l \leftarrow$ partial solution from root to $n_s$ in $\mathcal{T}$
    \State Add $(\mathbf{x}, \mathbf{y}^w, \mathbf{y}^l)$ to $\mathcal{P}$ 
\EndIf
\State $n_c \leftarrow$ randomly select a non-correct node at the same level of $n_w$  \Comment{Pairs in cousin nodes}
\If{$n_c \neq \emptyset$}
    \State $\mathbf{y}^l \leftarrow$ partial solution from root to $n_c$ in $\mathcal{T}$
    \State Add $(\mathbf{x}, \mathbf{y}^w, \mathbf{y}^l)$ to $\mathcal{P}$ 
\EndIf
\State $n_t \leftarrow$ randomly select a non-correct terminal node at the different level of $n_w$  \Comment{Pairs in non-same-level terminal nodes}
\If{$n_t \neq \emptyset$}
    \State $\mathbf{y}^l \leftarrow$ partial solution of $n_t$ in $\mathcal{T}$
    \State Add $(\mathbf{x}, \mathbf{y}^w, \mathbf{y}^l)$ to $\mathcal{P}$
\EndIf
\If{$n_s = n_c = n_t = \emptyset$}
    \State (Optional) Find $\mathbf{y}^l$ in another tree $\mathcal{T}^\prime$. \Comment{If no negative example found in all trees, all possible generated solutions are correct. No preference learning needed for this question.}
\EndIf
\State $n \leftarrow n_w$ \Comment{Go to next layer}
\EndWhile
\end{algorithmic}
\end{algorithm*}

\subsection{Policy-value model Details}
\label{sec:app_model}

\begin{figure}[ht]
\centering
  \includegraphics[width=0.85\columnwidth]{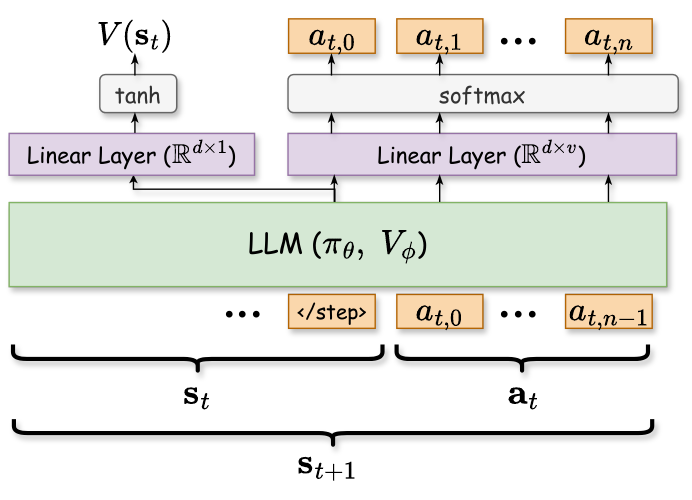}
  \caption{An overview of our policy-value model. $d$ represents the dimension of the hidden state in LLM, and $v$ represents the size of the vocabulary.}
  \label{fig:model}
\end{figure}

As shown in Figure~\ref{fig:model}, the value model $V_{\phi}$ and the LLM policy model $\pi_{\theta}$ are the same model but with different final layers. 
This design implies that these two models, $\pi_{\theta}$ and $V_{\phi}$, share the majority of their parameters. 
In practical implementation of the value loss, the value is only predicted on the last token of current reasoning step, representing the step-level preference.

\subsection{Experiment Environments}
All experiments were conducted on Ubuntu 22.04 equipped with 8 * NVIDIA A100 GPUs.
Our code mainly depends on Python 3.11 and PyTorch 2.2.1.
We use our customized \textit{Llama Factory}~\citep{zheng2024llamafactory} as the training framework and our customized \textit{vLLM}~\citep{KwonLZ0ZY0ZS23} as the inference framework\footnote{We have released our customized framework in our \href{https://github.com/MARIO-Math-Reasoning/Super_MARIO}{Github Repository}.}.
We trained all models with \textit{DeepSpeed ZeRO Stage2}~\citep{RajbhandariRRSH21} and \textit{Flash-Attention 2}~\citep{dao2023flashattention}.
The pre-trained LLMs are sourced from \textit{HuggingFace}\footnote{\url{https://huggingface.co}}.

\subsection{Prompt Example of our XML format}
\label{sec:app_xml_format}
To train the SFT model in executing mathematical reasoning, we utilize an XML format alongside zero-shot learning. 
This approach is adopted because the math-related pre-training corpora are predominantly harvested from the Internet, where HTML tags serve to distinguish various types of content, including text, equations, and code snippets.
In this work, each solution consists of both text analysis and code snippet, as shown in Figure~\ref{fig:example_xml_format}.

\begin{figure*}[h]
\begin{tcolorbox}[colback=wkyellow!50!white,colframe=wkyellow!80!orange,title=\textcolor{black}{An example of our SFT XML format:}, width=\textwidth]
\begin{small}

\begin{Verbatim}[commandchars=\\\{\}]
<question>Haley grows at the rate of 3 inches every year. If she is currently 20 inches tall,
what will be her height after 10 years?</question>
\textcolor{red}{<step>} 
\textcolor{red}{<p>} 
\textcolor{red}{To calculate Haley's height after 10 years, I need to add 10 times the growth rate of 3 inches to}
\textcolor{red}{her current height.}
\textcolor{red}{</p>}
\textcolor{red}{<code>}
\textcolor{red}{```python}
\textcolor{red}{current_height = 20}
\textcolor{red}{growth_rate = 3}
\textcolor{red}{years = 10}
\textcolor{red}{future_height = current_height + (growth_rate * years)}
\textcolor{red}{print(future_height)}
\textcolor{red}{```}
\textcolor{red}{</code>}
\textcolor{red}{<output>}
\textcolor{red}{50}
\textcolor{red}{</output>}
\textcolor{red}{</step>}
\textcolor{red}{<step>}
\textcolor{red}{<p>}
\textcolor{red}{I have calculated Haley's height after 10 years. Haley will be 50 inches tall}
\textcolor{red}{after 10 years.}
\textcolor{red}{</p>}
\textcolor{red}{<p>}
\textcolor{red}{Final Answer: $50$}
\textcolor{red}{</p>}
\textcolor{red}{</step>}
\end{Verbatim}
\end{small}
\end{tcolorbox}
\caption{An example of our SFT XML format. The text in black is prompt, and the text in \textcolor{red}{red} is model generation.}

\label{fig:example_xml_format}
\end{figure*}

\end{document}